\documentclass[letterpaper]{article} 
\usepackage[preprint]{aaai2027}  
\usepackage[hyphens]{url}  
\usepackage{graphicx} 
\usepackage{natbib}  
\usepackage{caption} 
\usepackage{amsmath}
\usepackage{amssymb}

\usepackage{booktabs}

\title{PeakBench: Benchmarking Resource-Aware\\
Tool Invocation in LLM Agents}
\author{
    Zhi-Kai Chen\textsuperscript{\rm 1,\rm 2},
    Xu-Xiang Zhong\textsuperscript{\rm 1,\rm 2},
    Song-Yan Li\textsuperscript{\rm 3},
    De-Chuan Zhan\textsuperscript{\rm 1,\rm 2},
    Han-Jia Ye\textsuperscript{\rm 1,\rm 2}\corresponding
}
\affiliations{
    \textsuperscript{\rm 1}School of Artificial Intelligence,
    Nanjing University, China\\
    \textsuperscript{\rm 2}National Key Laboratory for Novel Software Technology,
    Nanjing University, China\\
    \textsuperscript{\rm 3}Nanjing University, China\\
}

\begin{document}

\maketitle

\begin{abstract}
LLM agents increasingly solve tasks by invoking multiple tools, where parallel execution is essential for low latency but difficult to manage safely. Existing agent benchmarks primarily evaluate tool selection, argument generation, and end-to-end success under mostly serial execution, largely overlooking valid parallelization and resource-constrained scheduling. This missing scheduling dimension creates a practical failure mode: serial execution is safe but slow, while resource-agnostic parallel execution is fast but prone to avoidable resource overflows. To address this gap, we introduce PeakBench, a benchmark of executable multi-tool workflows with execution-grounded dependency annotations and measured resource profiles. A central challenge in evaluating such workflows is attribution: failures and inefficiencies may arise from incorrect dependency planning, poor resource-constrained scheduling, or both. PeakBench addresses this challenge with a two-part evaluation framework that disentangles logical planning from physical scheduling, with dedicated metrics for each dimension. Using this framework, we show that strong logical planning does not reliably translate into safe or efficient execution under resource constraints. We further show that exposing resource information can reduce avoidable overflows and improve resource utilization, making PeakBench a useful testbed for diagnosing resource-aware agent behavior. Code is available at \url{https://github.com/Czzzk/Staggering-the-Peaks}.
\end{abstract}


\section{Introduction}
Large Language Models (LLMs) have undergone a paradigm shift, transitioning from passive conversationalists to autonomous ``task-solvers'' capable of interacting with the physical and digital worlds~\cite{survey_agents,generative_agents,survey_rise}. At the heart of this evolution lies the mechanism of tool invocation, which enables agents to extend their reasoning capabilities through external APIs, databases, and computational engines.~\cite{toolformer,gorilla,react} As agents are increasingly deployed in complex, real-world workflows---ranging from automated software engineering~\cite{swe} to open-ended embodied or environment-interactive tasks~\cite{voyager}---the ability to interact with tools has become the definitive characteristic of ``agentic'' intelligence. 

\begin{figure}[t]
\centering
\includegraphics[width=\columnwidth]{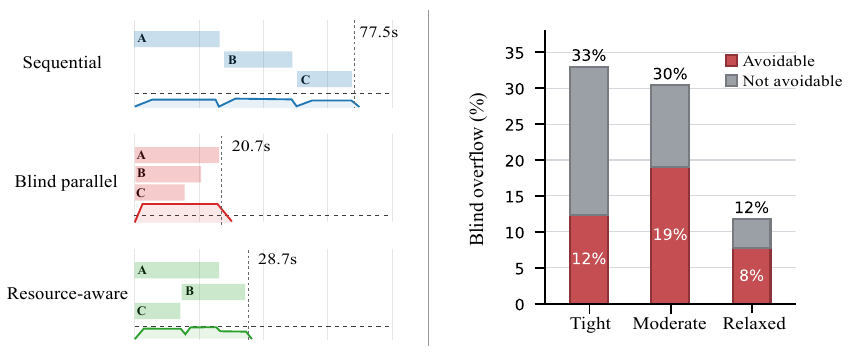}
\caption{Motivating tradeoff. Left: completion time and peak overflow for one workflow show that sequential execution is safe but slow, blind parallelism is fast but unsafe, and resource-aware scheduling preserves speedup without overflow. Right: blind-parallel overflows are decomposed across capacity profiles; avoidable overflow can be removed by a dependency-valid alternative schedule.}
\label{current_situation}
\end{figure}

\begin{figure*}[t]
\centering
\includegraphics[width=\textwidth]{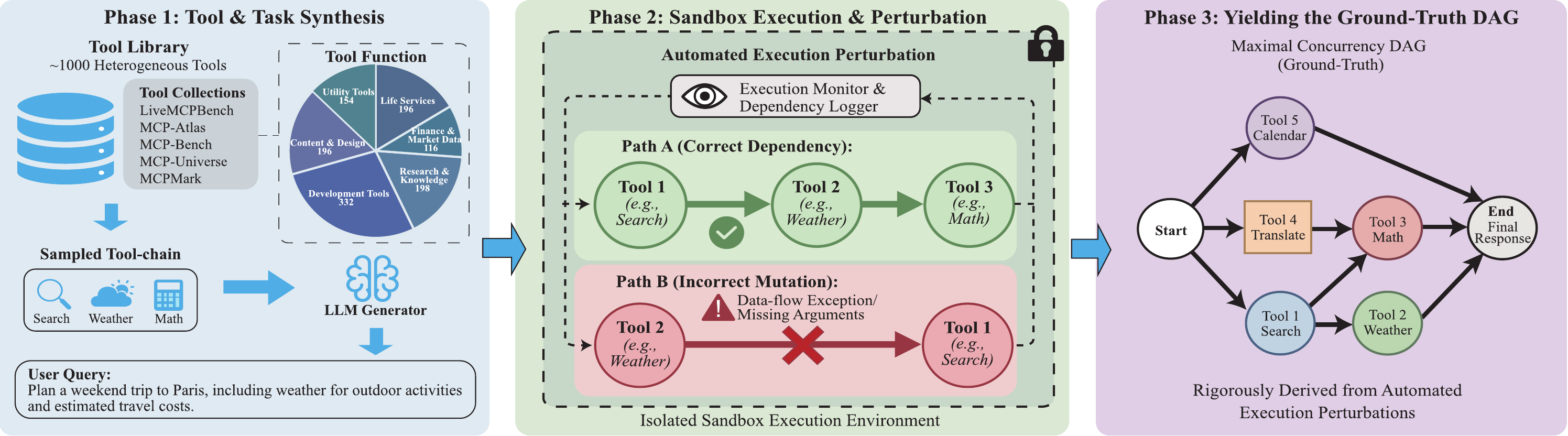}
\caption{
PeakBench dataset construction pipeline. Left: benchmark-seeded workflow synthesis samples MCP tools and generates executable multi-tool queries. Middle: sandbox execution observes step-level behavior and validates inter-step relations. Right: execution-order perturbation recovers prerequisite and concurrency structure used by both benchmark dimensions.}
\label{task_generation}
\end{figure*}

As task complexity escalates, modern agentic workflows demand concurrent execution to overcome the latency bottlenecks of sequential (step-by-step) models~\cite{cot,sot}. Safe concurrency first requires understanding step-level prerequisite and concurrency relations: which tool calls must wait for prior outputs, and which calls can run independently~\cite{got,llm_p}. Current agents may hallucinate false dependencies (needlessly serializing tasks and degrading throughput) or miss critical prerequisites (leading to execution-blocking errors)~\cite{plan_bench,cannot_correct}. Yet these logical relations only define what may run in parallel; they do not determine whether those parallel calls can safely share finite infrastructure.

Furthermore, even when agents identify or are given valid dependency and concurrency structure, a critical systemic vulnerability emerges. Current frameworks conflate ``logical independence'' with ``execution readiness.'' Because they are fundamentally resource-agnostic, they operate under the naive assumption of infinite infrastructure capacity~\cite{vllm,orca,deepspeed,aios}. Once independent tasks are identified, agents greedily dispatch all parallelizable tool invocations simultaneously without any physical scheduling awareness. As depicted in Figure~\ref{current_situation}, this unmanaged translation from logical parallelism to physical execution triggers massive ``Resource Bursts.'' Heavy, resource-intensive tools compete for finite hardware, leading to sharp spikes in infrastructure strain~\cite{alpa_serve,splitwise}, severe queuing delays, and catastrophic service outages---a systemic bottleneck we formalize as the ``peak load'' problem.


Despite the severity of these parallelization and physical-scheduling bottlenecks, the evaluation of LLM agents remains overwhelmingly ``accuracy-centric.'' Existing benchmarks~\cite{mcptoolbench++,api_bank,agentbench,webarena,gaia,big-bench}, such as ToolBench and APIBank, primarily evaluate tool selection, argument generation, and end-to-end success under mostly serial execution. While task success is a necessary condition, this narrow focus creates a critical research gap: these frameworks largely overlook valid parallelization and implicitly operate under the assumption of infinite and instantaneous resources. They ignore the temporal congestion and hardware footprint of tool-calling. In real-world deployments, an agent that reaches the correct answer but triggers a system-wide crash due to unmanaged request spikes is effectively unusable. Yet, current methodologies lack the vocabulary and metrics to quantify such operational failures.

End-to-end tool-agent execution makes failure attribution difficult. A slow or
failed workflow may reflect incorrect tool selection, invalid arguments, missing
dependencies, unnecessary serialization, unsafe parallelism, or resource
overload under a particular machine capacity. Treating these outcomes as a
single task-success score therefore obscures whether the agent failed to recover
the step-level dependency structure, failed to schedule otherwise valid tool
calls under resource constraints, or failed for unrelated tool-use reasons. This
motivates a decoupled benchmark design in which executable workflows provide a
validated task substrate, execution-grounded dependency and concurrency
relations isolate logical planning, and measured resource profiles make
physical scheduling observable.

Following this design, PeakBench constructs
executable multi-tool workflows, derives execution-grounded dependency and
concurrency annotations, and attaches empirically measured resource profiles to
tool invocations. Dimension~I evaluates logical planning by asking the agent to
recover which workflow steps are prerequisites and which can run concurrently.
Dimension~II evaluates physical scheduling by giving this verified structure
and asking the agent to assign execution timestamps under finite resource
budgets. This separation makes failures attributable: a model may fail because
it misunderstands dependencies, because it overloads resources, or because it
does both.

Our evaluations using PeakBench reveal a new failure mode: agents that appear
competent at logical workflow planning can be ``resource-blind'' when
translating the workflow into physical execution. To test whether this failure
is diagnosable, we include Resource-Aware Scheduling Context (RASC), a simple
baseline that exposes resource metadata before scheduling and tests whether
models can use such information to reduce avoidable overflow and improve
utilization. Our core contributions are summarized as follows:
\begin{itemize}
\item We identify logical planning and resource-constrained physical scheduling
as distinct evaluation targets often conflated in current LLM-agent benchmarks,
and formalize the ``peak load'' failure mode that arises when logically valid
parallel tool calls exceed finite resource capacity.
\item We introduce PeakBench, a decoupled benchmark with executable multi-tool
workflows, execution-grounded dependency and concurrency annotations, measured
resource profiles, and separate protocols for logical planning and physical
scheduling.
\item Across representative LLMs, we show that planning strength does not
reliably imply safe scheduling, while the RASC baseline yields measurable but
model-dependent gains when resource information is exposed.
\end{itemize}


\begin{figure*}[t!]
\centering
\includegraphics[width=\textwidth]{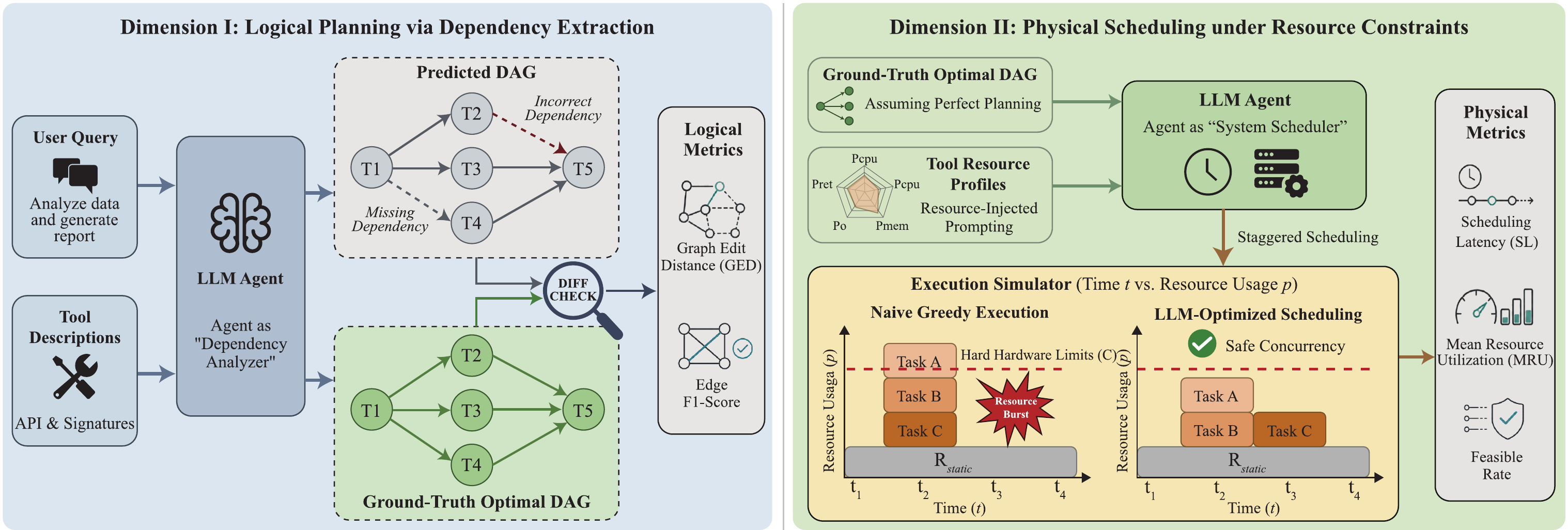}
\caption{PeakBench evaluation pipeline. Dimension~I evaluates the model as a ``Dependency Analyzer'' by comparing its predicted prerequisite structure against the execution-grounded structure using dependency metrics. Dimension~II gives the model the verified structure and evaluates it as a ``System Scheduler'' under measured resource profiles, using Scheduling Latency, Capacity Violation Area, and strict MRU.}
\label{evaluation_overview}
\end{figure*}

\section{Preliminary: Agentic Workflows as Constrained Scheduling Problems}

We view multi-tool agent execution as a constrained workflow execution problem.
Beyond selecting tools and producing a final answer, an agent must preserve
step-level data-flow validity while scheduling invocations under finite
infrastructure capacity. Let $\mathcal{V} = \{v_1, v_2, \dots, v_n\}$ denote
the tool invocations in a workflow and
$\mathcal{S} = \{t_1, t_2, \dots, t_n\}$ their activation schedule. A slow or
unsafe run may result from an invalid prerequisite structure, a poor
resource-constrained schedule in practice, or both.

\textbf{Overall Execution Objective.}
Given a benchmark task $x$, available tool descriptions, and candidate
invocations $\mathcal{V}$, the agent must produce an executable plan and assign
activation timestamps $\mathcal{S}$. The plan must preserve data-flow
prerequisites, and the schedule must keep concurrent resource demand within
machine capacity. Because the prerequisite structure is not assumed to be
given, it is implicit in whether the resulting execution can correctly use
intermediate outputs and complete the task. Resource-aware agentic execution
can therefore be formulated as:
\begin{equation}
\label{eq:overall_execution_objective}
\begin{aligned}
\min_{\mathcal{V},\mathcal{S}} \quad & \mathrm{SL}(\mathcal{S})
\quad \text{s.t.} \quad \Phi(x;\mathcal{V},\mathcal{S}) = 1, \\
& \mathbf{L}_{\mathcal{S}}(\tau) \preceq \mathbf{C},\quad \forall \tau .
\end{aligned}
\end{equation}
Here $\Phi(x;\mathcal{V},\mathcal{S})=1$ denotes logical execution success: the
selected invocations, arguments, and intermediate data dependencies are
sufficient to complete the task. The second constraint is physical execution
feasibility: $\mathbf{L}_{\mathcal{S}}(\tau)$ is the aggregate resource load
induced by $\mathcal{S}$ and must remain within capacity $\mathbf{C}$.
Equation~\ref{eq:overall_execution_objective} captures the central coupling:
successful agentic execution requires both a valid dependency structure and a
resource-feasible schedule. The equation intentionally leaves both constraints
abstract; the two stages below instantiate them for evaluation.

\textbf{Decoupled Evaluation.}
A monolithic end-to-end score cannot tell whether a violation of
Equation~\ref{eq:overall_execution_objective} comes from a wrong dependency
structure, a poor physical schedule, or both. PeakBench therefore turns the
two constraints in Equation~\ref{eq:overall_execution_objective} into two
separate benchmark dimensions. To do so, it derives an execution-grounded
prerequisite relation $\mathcal{R}^{\star}$ over workflow steps and evaluates:
\begin{equation}
\label{eq:decoupled_evaluation}
\begin{aligned}
\Phi(x;\mathcal{V},\mathcal{S}) = 1
&\;\Longrightarrow\;
\mathrm{Eval}_{\mathrm{plan}}(\hat{\mathcal{R}}, \mathcal{R}^{\star}), \\
\mathbf{L}_{\mathcal{S}}(\tau) \preceq \mathbf{C}
&\;\Longrightarrow\;
\mathrm{Eval}_{\mathrm{sched}}(\mathcal{S}; \mathcal{R}^{\star}, \mathbf{r}, \mathbf{C}) .
\end{aligned}
\end{equation}
Thus, Dimension~I evaluates the logical-success constraint by comparing the
predicted prerequisite structure with $\mathcal{R}^{\star}$, while
Dimension~II evaluates the resource-feasibility constraint by fixing
$\mathcal{R}^{\star}$ and replaying $\mathcal{S}$ under $\mathbf{r}$ and
$\mathbf{C}$. This decomposition preserves the structure of the overall
execution problem while making the source of failure attributable. We next
detail the two subproblems.

\textbf{Stage 1: Logical Planning (Dependency Extraction).}
Before execution, an agent must infer which tool calls can run concurrently and
which must be sequential. We formalize this as a prerequisite relation
$\mathcal{R} \subseteq \mathcal{V}\times\mathcal{V}$ over workflow steps. A
pair $(v_i,v_j)\in\mathcal{R}$ means that $v_i$ must complete before $v_j$ can
begin because $v_j$ depends on $v_i$'s output or side effect. The complement of
the transitive prerequisite relation determines which step pairs are validly
concurrent:
\begin{equation}
\label{eq:concurrency_relation}
\mathcal{C}(\mathcal{R})
=
\left\{
\{v_i,v_j\}\,:\,
(v_i,v_j)\notin\mathcal{R}^{+},\,
(v_j,v_i)\notin\mathcal{R}^{+}
\right\},
\end{equation}
where $\mathcal{R}^{+}$ denotes transitive closure. Let
$\hat{\mathcal{C}}=\mathcal{C}(\hat{\mathcal{R}})$ and
$\mathcal{C}^{\star}=\mathcal{C}(\mathcal{R}^{\star})$. PeakBench instantiates
the logical-success constraint as
$\mathrm{Eval}_{\mathrm{plan}}(\hat{\mathcal{R}},\mathcal{R}^{\star})$:
\begin{equation}
\label{eq:planning_metrics}
\left\langle
D_{\mathrm{dep}}(\hat{\mathcal{R}},\mathcal{R}^{\star}),
A_{\mathrm{conc}}(\hat{\mathcal{C}},\mathcal{C}^{\star})
\right\rangle .
\end{equation}
Here $D_{\mathrm{dep}}$ captures discrepancy in required prerequisite
relations, while $A_{\mathrm{conc}}$ captures agreement on which step pairs can
run concurrently.

\textbf{Stage 2: Physical Scheduling (Resource Allocation).}
While $\mathcal{R}$ dictates what can be parallelized, the infrastructure
capacity $\mathbf{C}$ dictates how much can be parallelized safely. Each
invocation $v_i$ has duration $d_i$ and measured resource footprint
$\mathbf{r}_i$, and any candidate schedule induces the resource-load vector:
\begin{equation}
\label{eq:resource_load}
\mathbf{L}(\tau) = \sum_{i=1}^{n} \mathbf{r}_i \, \mathbb{1} \left( \tau \in [t_i, t_i + d_i) \right)
\end{equation}
where $\mathbb{1}(\cdot)$ is an indicator function for active execution. Given
a prerequisite relation $\mathcal{R}$ and resource footprints
$\{\mathbf{r}_i\}_{i=1}^n$, a schedule must respect prerequisite order
($t_j\geq t_i+d_i$ for all $(v_i,v_j)\in\mathcal{R}$) while keeping aggregate
load within capacity. PeakBench instantiates the resource-feasibility
constraint as
$\mathrm{Eval}_{\mathrm{sched}}(\mathcal{S};\mathcal{R}^{\star},\mathbf{r},\mathbf{C})$:

\begin{equation}
\label{eq:scheduling_metrics}
\left\langle
T_{\mathrm{end}}(\mathcal{S}),
V_{\mathrm{cap}}(\mathcal{S};\mathbf{r},\mathbf{C}),
U_{\mathrm{safe}}(\mathcal{S};\mathbf{r},\mathbf{C})
\right\rangle .
\end{equation}
Here $T_{\mathrm{end}}$ captures completion time, $V_{\mathrm{cap}}$ captures
the severity of capacity breaches, and $U_{\mathrm{safe}}$ captures resource
utilization only when the schedule remains feasible. $\preceq$ denotes
element-wise comparison across resource dimensions.

\textbf{The ``Resource-Blind'' Bottleneck.}
A resource-blind scheduler observes prerequisite structure but not
$\mathbf{r}_i$ or $\mathbf{C}$. Even with correct logical structure, it may
therefore launch all ready steps as
early as possible and violate $\mathbf{L}(\tau) \preceq \mathbf{C}$, which
PeakBench isolates in Dimension~II.

\section{PeakBench}

PeakBench evaluates whether LLM agents can act as \emph{resource-aware
workflow executors}, not only functional tool selectors. It couples a large
catalog of MCP-compatible tools with semantically plausible, executable
multi-tool workflows whose dependencies and resource footprints are empirically
grounded. Each instance supports two linked evaluations: predicting the
execution-grounded dependency structure for logical planning, and scheduling
the same structure under resource constraints for physical execution.

\subsection{Benchmark Construction}
PeakBench operationalizes this design through sandbox-validated workflows,
execution-grounded step relations, empirical resource profiling, and a two-part
evaluation protocol.

\textbf{Tool Catalog and Resource Characterization.}
PeakBench builds on approximately 1.2K MCP-compatible tools spanning about 130
servers, aggregated from representative MCP-based tool-use ecosystems and
agentic benchmarks, including LiveMCPBench~\cite{live_mcp},
MCP-Atlas~\cite{mcp_atlas}, MCP-Bench~\cite{mcp_bench},
MCP-Universe~\cite{mcp_universe}, and MCPMark~\cite{mcp_mark}. The catalog
covers common agentic capabilities such as retrieval, code execution, file
manipulation, database access, web interaction, multimodal processing, and
model-in-the-loop computation.

We characterize these tools along functional and operational axes. The former
captures agentic roles and application domains, while the latter summarizes the
dominant bottlenecks shown in Table~\ref{tab:dual_taxonomy}. This distinction
matters because functionally similar tools can impose very different system
costs, so resource-aware scheduling cannot be inferred from tool semantics
alone. Each tool is further associated with an empirically measured resource
profile; the profiling protocol is detailed in
Appendix~\ref{appendix:resource_profiling}.

\begin{table}[t]
\centering
\small
\begin{tabular*}{\columnwidth}{@{\extracolsep{\fill}}lrlr@{}}
\toprule
\textbf{Functional} & \textbf{\#} & \textbf{Resource Tag} & \textbf{\#} \\
\midrule
Life Services & 196 & Lightweight & 436 \\
Finance \& Market Data & 116 & Memory-heavy & 632 \\
Research \& Knowledge & 198 & CPU-heavy & 201 \\
Development Tools & 332 & Network-heavy & 417 \\
Content \& Design & 196 & Disk-I/O-heavy & 64 \\
Utility Tools & 154 & Process-fanout & 7 \\
\bottomrule
\end{tabular*}
\caption{PeakBench tool catalog taxonomy. Functional categories are mutually exclusive; resource-cost tags are multi-label over profiling-eligible tools.}
\label{tab:dual_taxonomy}
\end{table}

\textbf{Benchmark-Seeded Workflow Synthesis.}
Workflow synthesis starts from task domains, question styles, and tool-usage
patterns observed in existing API/MCP benchmarks. Given these seeds, PeakBench
samples MCP tools and generates semantically nearby executable workflows over
the sampled components. This preserves benchmark-grounded task intent and
tool-composition patterns while allowing controlled variation in invocation
scale, dependency depth, parallel branch width, and resource heterogeneity.

We define difficulty through workflow structure rather than linguistic
complexity. PeakBench contains 300 executable workflows, stratified into 150
easy, 100 medium, and 50 hard tasks, with tier definitions reported in
Appendix~\ref{appendix:construction_details}. For each tier, we sample MCP
servers and tools from benchmark-derived domains, then prompt an LLM to
synthesize an executable user query that requires the selected components and
induces a concrete multi-step workflow. Figure~\ref{task_generation} (Left)
illustrates this benchmark-seeded synthesis stage.

\textbf{Execution-Grounded Step Relation Annotation.}
After a workflow query is synthesized, we execute its tool invocations inside a
controlled containerized sandbox exposing the sampled MCP tools and servers.
This stage is used to observe and annotate step-level relations rather than
merely to check whether the query is executable. We first collect execution
traces showing which steps consume prior outputs and which can proceed
independently. We then perturb candidate execution orders inside the sandbox and
monitor data-flow failures,
such as missing inputs or violated prerequisites. These observations identify
required prerequisite relations and feasible concurrent groups, producing the
execution-grounded structure used by both benchmark dimensions, as illustrated
in Figure~\ref{task_generation} (Middle, Right).

This execution-grounded structure plays a dual role in PeakBench's decoupled
evaluation paradigm. In the first dimension, it serves as the structural ground
truth against which an agent's predicted dependency structure is evaluated. In
the second dimension, it is provided directly to the agent as an oracle
workflow specification, thereby isolating resource-aware scheduling from
uncertainty in logical planning.

\subsection{Decoupled Evaluation Protocol}

Traditional agent benchmarks primarily evaluate whether a model can iteratively
select tools to reach a correct final answer. PeakBench instead evaluates
whether a validated multi-tool workflow can be executed efficiently under
finite infrastructure constraints. To avoid conflating workflow-understanding
errors with scheduling errors, we separate the evaluation into two dimensions:
Dimension~I tests dependency recovery, while Dimension~II fixes the
execution-grounded prerequisite structure and tests resource-aware scheduling. The following
subsections specify the inputs, outputs, and metrics for each dimension.

\textbf{Dimension I: Logical Planning via Dependency Extraction.}
The first evaluation dimension treats the agent as a
\emph{Dependency Analyzer}. Its objective is to test whether the model can recover
the intrinsic data-flow constraints of a multi-tool workflow before physical
execution. This capability is foundational: without a sound structure, later
parallelization or resource allocation would rest on flawed and unsafe
assumptions.

Given a user query together with the descriptions of the relevant tools and
servers, the agent must output a prerequisite structure indicating which tool
invocations must precede others and which can be safely executed in parallel.
For metric computation, this prerequisite structure is represented as a DAG.
To establish ground truth without exhaustive human annotation, we use the
execution-grounded structure recovered through sandbox perturbation, as
described above. This procedure yields an empirical structural target derived
from actual tool behavior rather than manual interpretation alone.

We compare the agent's predicted prerequisite structure against this
execution-derived target using two metrics. First, Graph Edit Distance (GED)
measures the minimum number of graph operations required to transform the
predicted structure into the ground-truth structure. Second, Edge F1 evaluates
the precision and recall of the predicted prerequisite edges, penalizing both
hallucinated dependencies and missing constraints. Together, these metrics
quantify whether the agent can recover the logical structure necessary to
maximize concurrency.

\textbf{Dimension II: Physical Scheduling under Resource Constraints.}
The second evaluation dimension treats the agent as a
\emph{System Scheduler}. While dependency extraction identifies the
\emph{potential} for concurrency, realizing that potential requires physical
orchestration under finite infrastructure budgets. This dimension therefore
evaluates whether an agent can flatten latency without triggering resource
contention, peak overload, or system-level failure.

Unlike Dimension~I, the goal here is not to infer workflow structure. Instead,
we directly provide the execution-grounded prerequisite structure as an oracle
input, thereby removing ambiguity about precedence constraints and isolating
the agent's scheduling ability. Given this verified structure together with
historical resource profiles, the agent must decide when each tool invocation
should be activated so as to balance parallel throughput against physical
feasibility.

To ground this task in system reality, we use the measured resource profiles
defined above and adopt a dual-state cost model to capture temporal resource
behavior. The total consumption for invocation $v$ with input $x$ is defined as
\begin{equation}
\label{eq:dual_state_resource}
\mathbf{r}_{\mathrm{total}}(v, x)=
\mathbf{r}_{\mathrm{static}}(v)+\mathbf{r}_{\mathrm{dynamic}}(v, x),
\end{equation}
where $\mathbf{r}_{\mathrm{static}}(v)$ represents the persistent baseline overhead
required to keep a tool ready (e.g., loaded model weights in GPU memory), and
$\mathbf{r}_{\mathrm{dynamic}}(v, x)$ captures the transient surge triggered during
active execution. We estimate these quantities through high-frequency system
telemetry collected across diverse instructions $\mathcal{X}_i$.

The agent must assign execution timestamps to minimize end-to-end latency while
preventing overlapping tool activations from breaching hard physical limits.
We evaluate schedules using three metrics: Scheduling Latency
(SL), Capacity Violation Area (CVA), and strict mean resource utilization
(strict MRU). Because resource dimensions have different physical units, we
first normalize each dimension by its machine capacity. For a replayed
schedule, let segment $k$ have duration $\Delta_k$ and dimensionless
utilization ratio $u_{k,m}=L_{k,m}/C_m$ on resource dimension $m$. We compute
\begin{equation}
\label{eq:cva}
\mathrm{CVA}=\sum_k \Delta_k \sum_m [u_{k,m}-1]_+,
\end{equation}
where $[x]_+=\max(x,0)$, so CVA is zero exactly when the schedule never exceeds
capacity in any dimension. Mean resource utilization (MRU) is the
duration-weighted average of the composite normalized load, and strict MRU
equals MRU only for zero-CVA schedules. Thus, CVA measures violation severity,
while strict MRU rewards high utilization only when the schedule is physically
safe. In this sense, Dimension~II evaluates not whether an agent understands
workflow logic, but whether it can translate a verified workflow structure into
a safe, efficient execution schedule.

\subsection{Human Quality Audit}

Although PeakBench derives its workflow structure through execution
perturbation, we also conduct a manual quality audit on approximately 100
sampled workflows. Annotators compare each generated structure against an
optimal structure derived from the query, tool descriptions, and execution
trace. About 94\% of the audited samples match the optimal structure. Since
each workflow typically requires about 4--5 minutes to audit manually, this
agreement supports the reliability of our
execution-grounded construction pipeline while illustrating why exhaustive
manual annotation is impractical.

\section{Resource-Aware Scheduling Context (RASC)}

The decoupled protocol exposes a natural diagnostic question: given the correct
workflow structure, can an agent use explicit resource information to schedule
more safely? Resource-Aware Scheduling Context (RASC) answers this question by
augmenting the scheduling input with pre-execution resource metadata. Before
execution, the agent sees not only which tool calls are ready, but also how
expensive those calls are and what capacity limits they must share. This lets
the model reason about physical contention before committing to an invocation
order.

Concretely, RASC takes four inputs: the user question, the verified workflow
structure, the target machine capacity $\mathbf{C}$, and a structured resource
profile for every tool invocation. Each profile records the estimated execution
duration and measured resource footprint of the invocation. The context also
states the execution semantics: tool calls
whose dependencies are satisfied may run concurrently, but the aggregate load
of all active calls should remain below the machine capacity. The model is
then asked to output an execution schedule, i.e., a start time or scheduling
delay for each workflow node.

Formally, for a workflow question $q$ and verified prerequisite structure
$\mathcal{R}^{\star}$, the resource-blind scheduler observes only the logical
planning context, whereas RASC additionally exposes the invocation profiles
$\mathcal{P}=\{(d_i,\mathbf{r}_i)\}$ and machine capacity:
\begin{equation}
\label{eq:rip_prompt_mapping}
\pi_{\theta}^{\mathrm{base}}=\pi_{\theta}(q,\mathcal{R}^{\star}), \qquad
\pi_{\theta}^{\mathrm{RASC}}=\pi_{\theta}(q,\mathcal{R}^{\star},\mathcal{P},\mathbf{C}).
\end{equation}
Thus, RASC changes only the scheduling context supplied to the model at decision
time, not the model parameters or the verified workflow structure.

This changes the agent's decision criterion from ``launch every ready tool as
early as possible'' to ``launch ready tools only when their combined resource
pressure is acceptable.'' RASC therefore targets peak load by preserving
parallelism where capacity permits while discouraging resource bursts. It
requires no retraining or external optimizer; the JSON-style context schema is
shown in Appendix~\ref{appendix:rip_prompt_format}.

\section{Experiment}

\subsection{Experimental Setup}

\textbf{Models.}
We evaluate eight frontier API models on PeakBench:
DeepSeek-V4-Flash, DeepSeek-V4-Pro, GLM-5, Kimi-K2.5, Claude Sonnet 4.6,
GPT-4.1, GPT-5, and o3. All models are accessed through their public API
interfaces and evaluated under the same prompting and parsing protocol.

\textbf{Benchmark and scheduling protocol.}
Both benchmark dimensions use the same PeakBench questions and
execution-grounded workflows. Dimension~I asks the model to recover the
dependency structure from the task and tool descriptions. Dimension~II gives the
model the verified structure and asks it to assign start timestamps to all tool
invocations under three machine profiles (\textit{small}, \textit{medium},
and \textit{large}); the profile construction is reported in
Appendix~\ref{appendix:machine_profiles}.

We compare two model settings. The no-profile setting provides the verified
structure but
hides tool-level resource telemetry. RASC additionally provides structured
resource profiles and machine capacities. We also include non-LLM baselines:
ASAP, serial topological execution, and resource-constrained list schedulers
ordered by duration, critical-path length, or normalized resource pressure.
The same simulator replays all schedules and computes the Dimension~II
metrics.

\textbf{Metrics.}
We use the Dimension~I and Dimension~II metrics defined above: GED and Edge F1
for dependency recovery, and Scheduling Latency (SL), Capacity Violation Area
(CVA), and strict mean resource utilization (strict MRU) for scheduling.
Arrows in the tables indicate whether lower or higher values are better.

\begin{table*}[t]
\centering
\small
\begin{tabular*}{\textwidth}{@{\extracolsep{\fill}}lccccc@{}}
\toprule
& \multicolumn{2}{c}{\shortstack{\textbf{Dimension I:}\\\textbf{Logical Planning}}}
& \multicolumn{3}{c}{\shortstack{\textbf{Dimension II:}\\\textbf{Scheduling Benchmark}}} \\
\cmidrule(lr){2-3} \cmidrule(lr){4-6}
\textbf{Model}
& GED $\downarrow$
& Edge F1 $\uparrow$
& SL (s) $\downarrow$
& CVA $\downarrow$
& Strict MRU $\uparrow$ \\
\midrule
GPT-5~\cite{openai_gpt5}                   & \textbf{0.42} & \textbf{0.839} & 13.28 & \underline{3.698} & \underline{0.125} \\
DeepSeek-V4-Pro~\cite{deepseek_v4}         & \underline{0.54} & \underline{0.807} & 12.63 & 4.032 & 0.120 \\
OpenAI o3~\cite{openai_o3}                 & 0.66 & 0.771 & 12.98 & 4.307 & \textbf{0.127} \\
GLM-5~\cite{glm5}                          & 0.65 & 0.772 & \underline{11.81} & 4.616 & 0.119 \\
Claude Sonnet 4.6~\cite{claude_sonnet46}   & 0.67 & 0.764 & 13.20 & 4.358 & 0.126 \\
DeepSeek-V4-Flash~\cite{deepseek_v4}       & 0.81 & 0.733 & 12.62 & \textbf{3.458} & 0.124 \\
Kimi-K2.5~\cite{kimi_k25}                  & 1.02 & 0.688 & \textbf{11.16} & 4.431 & 0.120 \\
GPT-4.1~\cite{openai_gpt41}                & 1.14 & 0.663 & 13.13 & 4.588 & 0.121 \\
\midrule
Baseline           & -- & -- & 8.62 & 5.865 & 0.080 \\
\bottomrule
\end{tabular*}
\caption{PeakBench benchmark results across logical planning and physical scheduling on the same PeakBench evaluation questions. Baseline denotes the earliest-ready execution state before model scheduling and is not ranked against model outputs. Bold and underline mark the best and second-best model values within each metric; ties are marked consistently. Dimension~II reports no-profile model scheduling before applying RASC.}
\label{tab:peakbench_api_results}
\end{table*}

\subsection{Benchmark Results}

Table~\ref{tab:peakbench_api_results} reports PeakBench's two benchmark
dimensions. The Baseline row gives the shared earliest-ready execution state
before model scheduling, while each model row reports the no-profile schedule
produced by that model. Dimension~II uses the verified structure and evaluates
Scheduling Latency, CVA, and strict MRU under resource limits before applying
RASC. Relative to the Baseline, models can reduce CVA and improve strict MRU
without resource telemetry, but mainly by adding conservative delays, which
raises Scheduling Latency. This reveals the resource-blind
scheduling regime that RASC later targets: models know the workflow structure,
but not which concurrent calls compete for the same resources.

The contrast between Dimension~I and Dimension~II shows that strong logical
planning does not imply strong resource-aware scheduling.
GPT-5 and DeepSeek-V4-Pro achieve the strongest dependency-extraction results,
yet their no-profile scheduling results remain close to weaker logical
planners under the physical metrics. PeakBench evaluates two
complementary capabilities: recovering the workflow structure and executing
that structure under finite resource budgets.

\begin{figure}[t]
\centering
\includegraphics[width=\columnwidth]{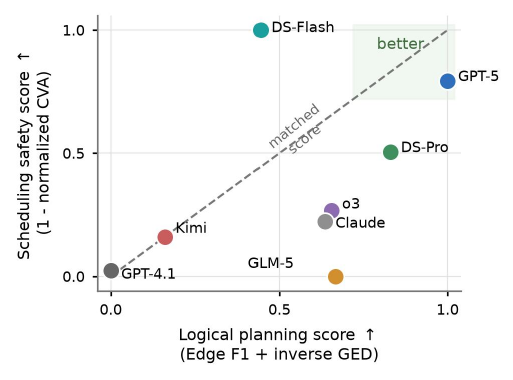}
\caption{Planning--scheduling decoupling across models. Both axes are normalized higher-better. The x-axis averages Edge F1 and inverse normalized GED; the y-axis uses one minus normalized CVA. The dashed diagonal indicates matched planning and scheduling scores.}
\label{fig:planning_scheduling_scatter}
\end{figure}

The case-level correlation analysis in
Appendix~\ref{appendix:planning_scheduling_correlation} further validates this
separation. Figure~\ref{fig:planning_scheduling_scatter} shows the same
decoupling at the model level: models do not concentrate near the upper-right
or along the matched-score diagonal. Across models, dependency-extraction success for a workflow only
weakly predicts whether the same workflow will be scheduled safely under finite
resources.

\subsection{RASC Results}

Providing resource information changes the scheduling behavior rather than only
the input surface. We evaluate this effect from two angles: against classical
scheduling rules, and against each model's no-profile schedule.

\begin{table}[t]
\centering
\small
\begin{tabular*}{\columnwidth}{@{\extracolsep{\fill}}lccc@{}}
\toprule
\textbf{Method} & \textbf{SL (s) $\downarrow$} & \textbf{CVA $\downarrow$} & \textbf{Strict MRU $\uparrow$} \\
\midrule
ASAP & \textbf{8.62} & 5.865 & 0.080 \\
Serial & 15.19 & \textbf{2.925} & 0.097 \\
Best Rule & 9.13 & \textbf{2.925} & 0.141 \\
RASC & \underline{9.11} & \underline{2.938} & \textbf{0.165} \\
\bottomrule
\end{tabular*}
\caption{RASC compared with classical scheduling baselines on Dimension~II, averaged over three machine profiles. Bold and underline mark the best and second-best values within each metric; ties are marked consistently.}
\label{tab:rip_vs_classical_schedulers}
\end{table}

We also compare RASC against classical scheduling rules, including
earliest-ready scheduling (ASAP), serial execution, and resource-constrained
list scheduling. Table~\ref{tab:rip_vs_classical_schedulers}
reports the best RASC setting using DeepSeek-V4-Pro, the strongest complete
model under RASC in our runs. ASAP is consistently fastest but incurs many
resource violations, while serial execution reduces CVA at a large latency cost.
RASC nearly matches the best rule-based scheduler in CVA while achieving higher
strict MRU, without external optimization. This comparison positions RASC as a
context-level diagnostic baseline: it does not replace hand-crafted schedulers,
but shows that capable LLMs can use explicit resource metadata when deciding
tool timing, or expose structured workflow and resource information to an
external scheduler.

\begin{table}[t]
\centering
\small
\begin{tabular}{lccc}
\toprule
\textbf{Model} & \textbf{$\Delta$SL (s) $\downarrow$} & \textbf{$\Delta$CVA $\downarrow$} & \textbf{$\Delta$MRU $\uparrow$} \\
\midrule
GPT-5             & \textbf{-4.31} & -0.534 & +0.040 \\
DeepSeek-V4-Pro   & -3.51 & \underline{-1.094} & \textbf{+0.045} \\
OpenAI o3         & -3.95 & -1.053 & +0.039 \\
GLM-5             & -2.67 & \textbf{-1.457} & +0.041 \\
Claude Sonnet 4.6 & -3.91 & -1.092 & +0.021 \\
DeepSeek-V4-Flash & -3.49 & +0.034 & \underline{+0.039} \\
Kimi-K2.5         & -2.04 & -0.305 & -0.014 \\
GPT-4.1           & \underline{-4.29} & -0.423 & -0.028 \\
\bottomrule
\end{tabular}
\caption{Effect of resource-aware scheduling context. Values are RASC minus no-profile scheduling, averaged over three machine profiles. For $\Delta$SL and $\Delta$CVA, lower is better; for $\Delta$MRU, higher is better. Bold and underline mark the best and second-best values within each metric.}
\label{tab:prompt_injection_effect}
\end{table}

Table~\ref{tab:prompt_injection_effect} isolates the RASC effect
by comparing RASC with each model's no-profile schedule. For most models, RASC
reduces Scheduling Latency while lowering CVA and improving strict MRU,
suggesting that resource profiles help replace conservative delays with targeted
staggering. DeepSeek-V4-Flash is the exception on CVA, where RASC lowers latency
and improves strict MRU but slightly increases violation area, indicating that
resource telemetry must still be translated into a valid temporal schedule.

Additional analyses in Appendix~\ref{appendix:additional_results} provide
case-level support for these results. Appendix Figure~\ref{fig:rip_machine_deltas}
shows that RASC gains are largest under tighter capacity profiles. The
model-level breakdown shows that GLM-5, DeepSeek-V4-Pro, Claude Sonnet 4.6,
and o3 obtain the largest CVA reductions, while DeepSeek-V4-Flash slightly
increases CVA. Appendix Tables~\ref{tab:planning_scheduling_correlation},
~\ref{tab:profile_noise_robustness}, and~\ref{tab:rip_input_ablation} show that
logical planning success only weakly predicts physical scheduling safety, RASC
remains beneficial under noisy resource profiles, and the gain comes primarily
from CPU and memory profiles rather than generic latency awareness.

\textbf{Main findings.}
Taken together, Table~\ref{tab:peakbench_api_results},
Table~\ref{tab:rip_vs_classical_schedulers},
Table~\ref{tab:prompt_injection_effect}, and the analyses in
Appendix~\ref{appendix:additional_results} support four main findings. First,
strong logical planning does not imply strong
resource-aware scheduling. Second, no-profile models are not merely
earliest-ready schedulers: they often add conservative delays that reduce CVA
and improve strict MRU, but at substantially higher Scheduling Latency. Third,
RASC turns this coarse caution into more targeted staggering, lowering latency
while usually reducing CVA and improving strict MRU. Fourth, RASC nearly matches
the best rule-based scheduler in CVA while achieving higher strict MRU, but the
benefit remains model-dependent, showing that resource-aware orchestration is a
distinct capability rather than an automatic consequence of exposing resource
metadata.

\FloatBarrier

\section{Conclusion}

We introduced PeakBench, a decoupled benchmark for evaluating LLM agents'
ability to recover workflow dependencies and schedule tool
invocations under finite infrastructure constraints. By separating logical
planning from physical scheduling, PeakBench makes peak-load failures
measurable and attributable beyond accuracy-centric tool-use evaluation. Our
experiments show that RASC reduces capacity violations and overload, while
models still vary in how they translate resource profiles into
efficient execution plans.

\bibliography{aaai2027}

\clearpage
\appendix
\setcounter{secnumdepth}{2}

\section{Additional Related Work}
\label{appendix:related_work}

\textbf{LLM Agents and Tool-Augmented Reasoning.}
Tool-augmented LLM agents interleave reasoning with external actions through
APIs, environments, and specialized tools. ReAct~\cite{react} and
Toolformer~\cite{toolformer} established tool use as a core agent capability,
while systems such as Voyager~\cite{voyager} and SWE-bench~\cite{swe} show that
agents increasingly execute multi-step workflows rather than only generate
text. These works demonstrate tool-use capability, but they do not evaluate
whether concurrent tool invocations are physically scheduled under finite
resources.

\textbf{Evaluation Frameworks for LLM Agents.}
Agent benchmarks such as API-Bank~\cite{api_bank}, GAIA~\cite{gaia},
BIG-bench~\cite{big-bench}, LiveMCPBench~\cite{live_mcp},
MCP-bench~\cite{mcp_bench}, MCP-Universe~\cite{mcp_universe},
MCP-Atlas~\cite{mcp_atlas}, MCPToolBench++~\cite{mcptoolbench++}, and
MCPMark~\cite{mcp_mark} evaluate API use, tool selection, and task completion
over increasingly broad tool ecosystems. TPS-Bench~\cite{tps_bench} is closest
to our setting because it studies whether agents can plan and schedule tool
calls efficiently in compounding tasks, mainly through task success and
execution-efficiency measures such as time, tool-call turns, and cost.
PeakBench instead isolates physical scheduling under measured resource
profiles, evaluating whether logically valid parallel tool invocations violate
finite infrastructure capacity. More broadly, existing benchmarks are essential
for measuring functional capability and tool-orchestration efficiency, but they
largely abstract away resource contention.

\textbf{Constraint- and Cost-Aware Agent Evaluation.}
Recent benchmarks also study whether agents can satisfy constraints or optimize
costs during tool use. CCTU~\cite{cctu} evaluates tool use under explicit
constraints spanning resource, behavior, toolset, and response categories,
COMPASS~\cite{compass} studies constrained preference optimization in
multi-turn travel planning, and CostBench~\cite{costbench} evaluates
cost-optimal planning and adaptation in dynamic tool-use environments.
R-ConstraintBench~\cite{r_constraintbench} similarly stresses reasoning under
interacting planning and allocation constraints. PeakBench is complementary:
instead of treating resources primarily as semantic task constraints or monetary
costs, it attaches empirical infrastructure profiles to tool invocations and
measures whether the resulting concurrent execution exceeds finite capacities.

\textbf{Resource Efficiency and System Scheduling in AI.}
Classical scheduling and modern AI serving systems study how to allocate
limited compute resources. Systems such as vLLM~\cite{vllm} and
DeepSpeed-Inference~\cite{deepspeed} optimize inference throughput and memory
management, while AIOS~\cite{aios} explores operating-system abstractions for
LLM agents. Workflow-level systems such as LLM-as-Scheduler~\cite{llm_as_scheduler}
dynamically route queries among alternative agent workflows to reduce latency
and token cost. These works optimize infrastructure or workflow choice once
requests arrive, whereas PeakBench asks whether the agent-side workflow itself
exposes enough dependency and resource information to avoid creating avoidable
peak load.

\section{Detailed Resource Characterization of Agentic Tools} 
\label{appendix:resource_profiling}

To accurately quantify the infrastructure strain caused by agentic tool-use, we characterize each tool invocation through a structured 5-dimensional resource footprint: $\mathbf{r} = \langle \rho_{\mathrm{cpu}}, \rho_{\mathrm{mem}}, \rho_{\mathrm{gpu}}, \rho_{\mathrm{net}}, \rho_{\mathrm{io}} \rangle$. These dimensions are strategically selected to cover the complete hardware stack utilized by modern LLM-driven agents:

\textbf{General-Purpose Computing (CPU \& RAM).} We monitor CPU core utilization ($\rho_{\mathrm{cpu}}$) and memory residency ($\rho_{\mathrm{mem}}$) to capture the baseline algorithmic overhead. These dimensions represent the primary costs of general logic execution and the volatile storage required for tool runtimes and intermediate data structures.

\textbf{Specialized Hardware Acceleration (GPU \& VRAM).} Recognizing the prevalence of ``model-in-the-loop'' tools, we explicitly track GPU/VRAM allocation ($\rho_{\mathrm{gpu}}$). Unlike general-purpose memory, VRAM is a highly scarce resource in agentic clusters; tracking this dimension is critical for identifying bottlenecks in AI-intensive tasks such as image generation, local tensor operations, or specialized embedding retrievals.

\textbf{Connectivity and Persistence (Network \& Disk I/O).} We measure network throughput ($\rho_{\mathrm{net}}$) and disk I/O intensity ($\rho_{\mathrm{io}}$) to account for tools that are ``environment-interactive.'' This includes data-intensive operations such as large-scale web scraping, external API communications, or heavy read/write tasks in database management.

\textbf{Resource Profile Measurement Protocol.}
For each profiling-eligible tool, we generate three semantically different
valid inputs with an LLM, conditioned on the tool schema and description. This
input set is intended to cover typical argument patterns for the tool rather
than a single hand-picked example. We execute the tool once per generated input
under an instrumented sandbox and record wall-clock duration together with
CPU, memory, GPU, network, and disk-I/O telemetry.

We explicitly distinguish cold-start and warm-start profiles. A cold-start run
starts from an unloaded tool or server state and therefore captures persistent
setup costs such as process initialization, model loading, cache creation, or
connection setup. A warm-start run reuses the initialized tool state and
captures the transient cost of active invocation after the tool is ready. The
resulting profile stores both components: $\mathbf{r}_{\mathrm{static}}$ for
the persistent baseline footprint and $\mathbf{r}_{\mathrm{dynamic}}$ for the
per-invocation surge. Unless otherwise stated, scheduling simulations use the
aggregated profile obtained from the three inputs, while retaining the
cold-/warm-start distinction for tools whose setup cost materially affects
capacity pressure.

\section{Construction Details of PeakBench} 
\label{appendix:construction_details}

PeakBench is constructed to stress both dependency reasoning and physical
scheduling. We do not claim that the workflow set reproduces the full
distribution of real user requests. Instead, we use existing API/MCP agent
benchmarks to seed task domains and tool-use patterns, then synthesize nearby
executable workflows whose structure can be controlled and verified.
As summarized in Table~\ref{tab:benchmark_statistics}, the workflow set is
stratified by difficulty. Easy cases contain short, shallow workflows over a
small number of servers, while medium and hard cases increase both the number
of servers and the number of tool invocations. This design prevents the
benchmark from being dominated by either trivial single-server calls or
unrealistically large workflows, and creates a controlled progression from
simple dependency extraction to multi-server scheduling pressure.

\begin{table}[t]
\centering
\small
\begin{tabular}{lccc}
\toprule
\textbf{Difficulty} & \textbf{\# Servers} & \textbf{\# Invocations} & \textbf{\# Workflows} \\
\midrule
Easy   & 1--2   & 2--4    & 150 \\
Medium & 3--5   & 5--8    & 100 \\
Hard   & 6--10  & 10--15  & 50 \\
\midrule
Total  & --     & --      & 300 \\
\bottomrule
\end{tabular}
\caption{PeakBench workflow composition by difficulty tier.}
\label{tab:benchmark_statistics}
\end{table}

Table~\ref{tab:dual_taxonomy} describes the tool catalog used to instantiate
these workflows. The functional taxonomy ensures coverage across common agent
use cases such as development, research, finance, content generation, and
utility services. The resource-cost taxonomy is deliberately multi-label:
tools can be memory-heavy, CPU-heavy, network-heavy, or disk-I/O-heavy at the
same time. This matters for Dimension~II because scheduling failures often
come from overlapping heterogeneous costs rather than from a single tool type.
Together, Tables~\ref{tab:benchmark_statistics} and~\ref{tab:dual_taxonomy}
show that PeakBench combines controlled workflow complexity with diverse
resource profiles.

\section{Simulated Machine Profiles}
\label{appendix:machine_profiles}

Dimension~II evaluates schedules under three resource regimes derived from the
validated PeakBench workflows. For each resource, the small profile uses the maximum
of the 95th percentile single-step load and the 65th percentile earliest-layer
load; medium uses step p98 and layer p80; large uses step p99 and layer p95.
This construction creates progressively relaxed capacities while keeping all
profiles grounded in observed MCP tool costs.

\begin{table}[t]
\centering
\small
\begin{tabular}{lcc}
\toprule
\textbf{Profile} & \textbf{Capacity rule} & \textbf{No-prof. CVA} \\
\midrule
Small  & max(step p95, layer p65) & 11.924 \\
Medium & max(step p98, layer p80) & 5.068 \\
Large  & max(step p99, layer p95) & 0.602 \\
\bottomrule
\end{tabular}
\caption{Machine profiles used in Dimension~II simulation. Capacities are derived from observed single-step and earliest-layer loads, not fixed hardware specifications.}
\label{tab:machine_profiles}
\end{table}

\section{RASC Scheduling Context Format}
\label{appendix:rip_prompt_format}

RASC exposes the scheduling objective and resource metadata to the model in a
structured JSON-style block. The following template summarizes the fields used
in the scheduling context; concrete benchmark instances fill in the question, steps,
dependency structure, machine profile, and per-step resource measurements.

\begin{small}
\begin{verbatim}
{
  "task": "<question>",
  "steps": {
    "<step_id>": "<tool call>"
  },
  "dependency_structure": {
    "<step_id>": ["<prereq>", "..."]
  },
  "earliest_depth": {
    "<step_id>": "<layer>"
  },
  "scheduling_objective": {
    "primary": "obey dependencies",
    "capacity_constraint":
      "stay within capacity",
    "latency_constraint":
      "minimize safe makespan",
    "execution_semantics":
      "same-depth steps are concurrent"
  },
  "machine": {
    "cpu_capacity": "<cpu cores>",
    "memory_mb_capacity": "<memory MB>"
  },
  "resource_costs": {
    "schema_version": "resource_cost_v1",
    "steps": {
      "<step_id>": {
        "tool": "<tool_name>",
        "server": "<server_name>",
        "duration_s": "<duration>",
        "cpu_avg_cores": "<avg CPU>",
        "cpu_peak_cores": "<peak CPU>",
        "memory_mb": "<peak memory>",
        "relative_to_case": {
          "duration_norm": "<duration>",
          "cpu_peak_norm": "<CPU>",
          "memory_norm": "<memory>",
          "pressure_norm": "<pressure>"
        },
        "relative_to_machine": {
          "cpu_capacity_ratio": "<CPU/cap>",
          "mem_capacity_ratio": "<mem/cap>",
          "fits_cpu": "<bool>",
          "fits_memory": "<bool>"
        },
        "classes": ["<class>", "..."]
      }
    }
  },
  "required_output": {
    "requested_delay": {
      "<step_id>": "<delay>"
    },
    "reasoning_summary": ["short rationale"]
  }
}
\end{verbatim}
\end{small}

\section{Additional Experimental Results}
\label{appendix:additional_results}

\subsection{Capacity-Profile Sensitivity}

\begin{figure}[t]
\centering
\includegraphics[width=\columnwidth]{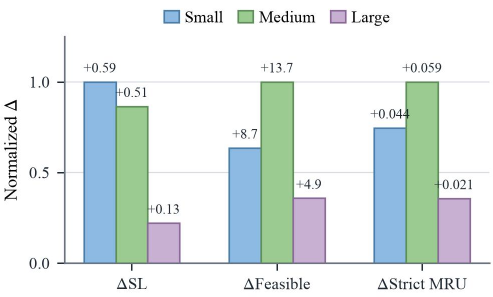}
\caption{Effect of RASC across machine profiles. Each group reports the change from no-profile scheduling to RASC; text labels show raw $\Delta$ values.}
\label{fig:rip_machine_deltas}
\end{figure}

Figure~\ref{fig:rip_machine_deltas} breaks down the effect of RASC by
machine profile. The gains are largest on the small and medium profiles, where
resource conflicts are more common, and smaller on the large profile, where
many workflows are already feasible. This supports the interpretation that RASC
mainly helps when dependency-valid schedules still need physical resource
coordination.

\subsection{Model-Level RASC Gains}

Averaged over machine profiles, RASC reduces CVA most strongly for GLM-5
(-1.457), DeepSeek-V4-Pro (-1.094), Claude Sonnet 4.6 (-1.092), and o3
(-1.053), while DeepSeek-V4-Flash shows a slight CVA increase.

\subsection{Relationship between Logical Planning and Scheduling}
\label{appendix:planning_scheduling_correlation}

Table~\ref{tab:planning_scheduling_correlation} reports per-model, case-level
correlations between Dimension~I planning outcomes and Dimension~II scheduling
metrics for the no-profile schedules in Table~\ref{tab:peakbench_api_results}. CVA and
strict MRU are averaged over the three machine profiles for the same case. The
weak correlations indicate that a model's success on dependency extraction for
a specific workflow does not reliably predict whether it will produce a safe and
efficient physical schedule for that workflow.

\begin{table*}[t]
\centering
\small
\begin{tabular*}{\textwidth}{@{\extracolsep{\fill}}lcccc@{}}
\toprule
\textbf{Model}
& \multicolumn{2}{c}{\textbf{CVA $\downarrow$}}
& \multicolumn{2}{c}{\textbf{strict MRU $\uparrow$}} \\
\cmidrule(lr){2-3} \cmidrule(lr){4-5}
& \textbf{Exact $r$} & \textbf{Edge F1 $r$}
& \textbf{Exact $r$} & \textbf{Edge F1 $r$} \\
\midrule
GPT-5             & -0.097 & -0.003 & +0.049 & +0.027 \\
DeepSeek-V4-Pro   & -0.115 & -0.023 & +0.094 & +0.067 \\
OpenAI o3         & -0.102 & -0.028 & +0.030 & +0.001 \\
GLM-5             & -0.098 & -0.036 & +0.098 & +0.102 \\
Claude Sonnet 4.6 & -0.063 & -0.000 & +0.054 & +0.062 \\
DeepSeek-V4-Flash & -0.093 & -0.022 & +0.036 & +0.031 \\
Kimi-K2.5         & -0.105 & -0.045 & +0.053 & +0.047 \\
GPT-4.1           & -0.053 & -0.004 & -0.052 & -0.063 \\
\bottomrule
\end{tabular*}
\caption{Per-model case-level Pearson correlation between logical planning and physical scheduling. CVA and strict MRU are averaged over three machine profiles for each case.}
\label{tab:planning_scheduling_correlation}
\end{table*}

\subsection{Resource-Profile Noise Robustness}

Table~\ref{tab:profile_noise_robustness} reports robustness to noisy resource
profiles. RASC remains beneficial under 10--50\% perturbations, but the
feasible-rate gain decreases as profile noise increases. This analysis uses the
seven model runs with complete noise-robustness results. Feasible rate is used
here as an auxiliary binary safety diagnostic; the main text reports CVA and
strict MRU as the primary Dimension~II metrics.

\begin{table*}[t]
\centering
\small
\begin{tabular*}{\textwidth}{@{\extracolsep{\fill}}llccc@{}}
\toprule
\textbf{Machine} & \textbf{Noise} & \textbf{Feasible $\Delta$} & \textbf{Violation reduction} & \textbf{SL overhead} \\
\midrule
Large  & 10\% & +4.4 & 0.279 & +0.14s \\
Large  & 25\% & +4.1 & 0.305 & +0.14s \\
Large  & 50\% & +3.2 & 0.354 & +0.14s \\
Medium & 10\% & +10.4 & 2.488 & +0.49s \\
Medium & 25\% & +8.4 & 2.601 & +0.50s \\
Medium & 50\% & +6.0 & 2.792 & +0.50s \\
Small  & 10\% & +6.2 & 4.536 & +0.59s \\
Small  & 25\% & +4.8 & 4.804 & +0.60s \\
Small  & 50\% & +3.5 & 5.282 & +0.60s \\
\bottomrule
\end{tabular*}
\caption{Resource-profile noise robustness. We perturb resource profiles before scheduling and report averages over seven complete model runs. Feasible $\Delta$ is the RASC improvement over no-profile scheduling.}
\label{tab:profile_noise_robustness}
\end{table*}

\subsection{RASC Input Ablation}

Table~\ref{tab:rip_input_ablation} isolates which parts of the RASC input are
responsible for the scheduling improvement. The main phenomenon is that
latency awareness alone is not enough: providing only duration information
does not improve feasibility over the no-profile setting. The improvement
appears once CPU and memory profiles are exposed, which indicates that the
method works by helping the model reason about capacity conflicts rather than
by simply encouraging shorter or longer schedules. We report feasible rate here
as a compact auxiliary safety diagnostic for this ablation.

\begin{table}[t]
\centering
\small
\begin{tabular}{lccc}
\toprule
\textbf{Input} & \textbf{Feas.} & \textbf{$\Delta$} & \textbf{Viol. red.} \\
\midrule
No profile & 74.9\% & -- & -- \\
Duration only & 74.9\% & +0.0 & $-0.0$\% \\
CPU+memory & 87.6\% & +12.7 & 48.5\% \\
Full profile & 87.8\% & +12.9 & 48.1\% \\
Full + ratio & 87.6\% & +12.7 & 45.8\% \\
\bottomrule
\end{tabular}
\caption{RASC input ablation using DeepSeek-V4-Flash on the PeakBench evaluation set and three machine profiles.}
\label{tab:rip_input_ablation}
\end{table}

The remaining rows in Table~\ref{tab:rip_input_ablation} further show that
CPU and memory account for nearly all of the observed gain in the current
benchmark. Adding the full profile or machine-normalized ratios changes the
result only marginally, suggesting that the dominant scheduling bottleneck is
the coarse capacity conflict captured by CPU and memory pressure. We therefore
use the full structured RASC format in the main experiments for completeness,
while interpreting its effect primarily as resource-capacity awareness rather
than generic context enrichment.


\end{document}